\documentclass[letterpaper, 10 pt, conference]{ieeeconf}

\usepackage{cite}
\usepackage{amsmath,amssymb,amsfonts}
\usepackage{algorithmic}
\usepackage{graphicx}
\usepackage{textcomp}
\usepackage{xcolor}
\usepackage{hyperref}
\usepackage{float}
\usepackage{tabularx}
\usepackage{makecell}

\IEEEoverridecommandlockouts

\title{\LARGE \bf
Preserving Sense of Agency: User Preferences for \newline Robot Autonomy and User Control across Household Tasks \thanks{This project was supported by the National Institute of Biomedical Imaging and Bioengineering (NIBIB) grant \#1R01EB034580-01 ``NRI: Adaptive Teleoperation Interfaces for In-Home Assistive Robots" and the National Science Foundation (NSF) grant \#2313998 ``Computer and Information Science and Engineering Graduate Fellowship".}}

\author{Claire Yang$^{1}$, Heer Patel$^{1}$, Max Kleiman-Weiner$^{1}$, Maya Cakmak$^{1}$
\thanks{$^{1}$ University of Washington, Seattle, WA, USA}
}

\begin{document}

\maketitle
\thispagestyle{empty}
\pagestyle{empty}

\begin{abstract}

Roboticists often design with the assumption that assistive robots should be fully autonomous. However, it remains unclear whether users prefer highly autonomous robots, as prior work in assistive robotics suggests otherwise. High robot autonomy can reduce the user's sense of agency, which represents feeling in control of one's environment. How much control do users, in fact, want over the actions of robots used for in-home assistance? We investigate how robot autonomy levels affect users' sense of agency and the autonomy level they prefer in contexts with varying risks. Our study asked participants to rate their sense of agency as robot users across four distinct autonomy levels and ranked their robot preferences with respect to various household tasks. Our findings revealed that participants' sense of agency was primarily influenced by two factors: (1) whether the robot acts autonomously, and (2) whether a third party is involved in the robot's programming or operation. Notably, an end-user programmed robot highly preserved users' sense of agency, even though it acts autonomously. However, in high-risk settings, e.g., preparing a snack for a child with allergies, they preferred robots that prioritized their control significantly more. Additional contextual factors, such as trust in a third party operator, also shaped their preferences.

\end{abstract}

\section{INTRODUCTION}

Household robots are intended to assist people with a variety of tasks, ranging from daily chores to caring for loved ones. People believe that robotic assistance will improve their lives and are positive about its potential to help them \cite{10.1145/1349822.1349827, rheman_longitudinal_2024, smarr_domestic_2014}. Despite generally favorable public
opinion, the effect of robotic assistance on users’ sense of control of their own and loved ones’ lives remains unclear. As roboticists aim to develop increasingly autonomous robots for the household, it is necessary to examine how a highly autonomous robot impacts the user. Robots that have high autonomy to make decisions and operate in a user's home potentially reduce users' sense of control, or \textit{agency}, even in `paternalistic' ways. Ghosh et al. \cite{ghosh_problem_2024} define this as ``robot-mediated paternalism,'' whereby a robot acts in ways intended to potentially benefit the user, but the user may be unaware of or not agree with the objective of that assistance. 

This is especially problematic in the context of assistive robots operating in user's homes. It has been shown that a greater sense of control is linked to improved physical health and psychological well-being in older adults \cite{hong_positive_2021}. A robot that reduces a user's sense of control in an everyday household context may actually worsen the user's well-being over time. This paper examines users' sense of agency over a robot assisting them in their home. We define \textit{sense of agency} as one's feeling of control over one's actions and the resulting consequences \cite{moore_what_2016}. Sense of agency can be affected by interactions with other agents, humans or machines \cite{wen_sense_2022}. Specifically, as a machine's autonomy increases, a user's sense of agency over the machine can be weakened \cite{berberian_automation_2012}. Therefore, it is clear that the user's sense of agency over a robot represents an important design consideration for assistive robots.

This work investigates how a robot user's sense of agency is affected by differing levels of robot autonomy. Given the potential use of household robots for varied tasks, we also aim to identify how sense of agency and user preferences for robot autonomy change with task context. Previous work showed that users prefer higher levels of robot autonomy when the task is simple and lower levels when the task is complex \cite{olatunji_levels_2022}. We examine how their preferences change with respect to the task's \textit{risk}, rather than its \textit{complexity}: 
a robot may complete low complexity tasks that contain high risks, such as organizing medicine bottles, and higher risk situations could lead to a greater desire for control to mitigate adverse consequences. Thus, 
we posit that \textit{risk has a more direct relationship with sense of agency than task complexity. }

Lastly, prior work has not investigated the impact of third-party involvement in operating the robot on user agency and preferences. Remote teleoperation is becoming increasingly popular for deploying robots to households, and companies are promising to do so in the next year \cite{torch_invasion_2025, humans_behind_robots_article}.
We contribute to this gap in the literature by investigating how and whether a user's sense of agency is a factor in their decision to prefer certain robots (including third-party teleoperated robots) over others with respect to the context's risks.

To address these questions, we conducted a two-part survey. The first part asked the participants to rate their sense of agency across four robot autonomy levels (fully autonomous, end-user programmed, third-party teleoperated, fully user-controlled). The second part presented vignettes of household scenarios with varying risks and asked users to rank and explain their preferences for the different robot autonomy levels. Based on responses, we develop a model showing that a user's sense of agency highly depends on third-party involvement and trust. Using this model, we find that a user's sense of agency can be preserved \textit{if the robot is programmed by the user, instead of relying on a third party to control the robot}. We also find that participants' robot autonomy preferences depended on the task risk and their trust in a third party operator being involved in the task.

\begin{figure}[t]
    \centering
    \includegraphics[width=\linewidth]{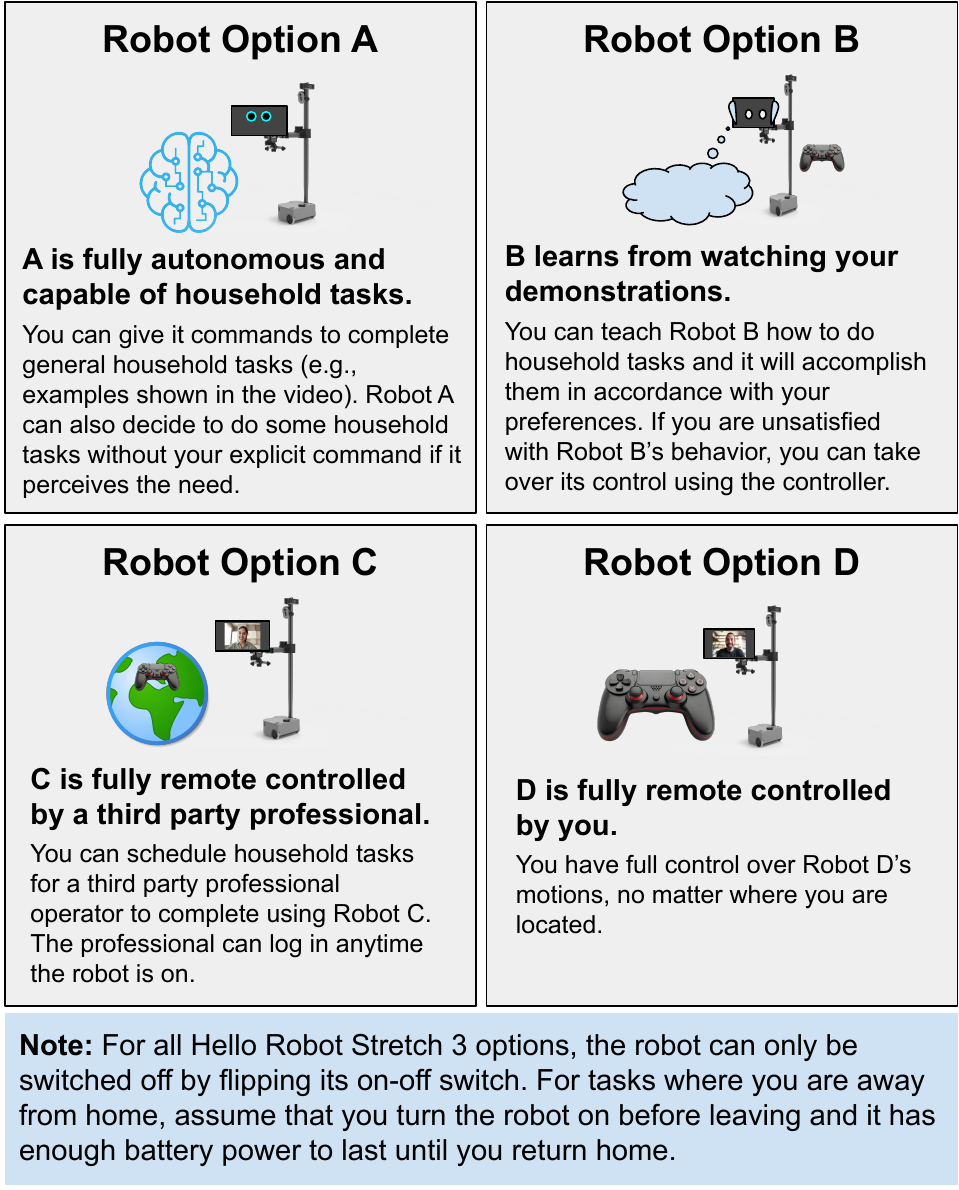}
    \caption{Robot types presented in the survey. (A) Fully autonomous, (B) end-user programmed, (C) third-party teleoperated, and (D) fully user-controlled.}
    \label{fig: robot type}
\end{figure}

\section{RELATED WORK}
Human's sense of agency and its connection to the autonomy of a robot has a rich background spanning multiple fields, including psychology, cognitive science, human-computer interaction (HCI), and human-robot interaction (HRI). Social psychologists and cognitive scientists have modeled human agency as a casual relationship between people's ability to decide, act, and account for the resulting consequences \cite{moore_what_2016, gallagher_philosophical_2000}. Broadly, the importance of human agency has been related to issues like freedom and self-determinism \cite{bandura_human_1989}. 

\subsection{Measuring Sense of Agency}
Methods measuring sense of agency can be roughly divided into two categories: implicit or explicit \cite{bennett_how_2023}. The two dominant implicit methods for measuring sense of agency are the comparator model, which calculates small changes in motor control \cite{frith_abnormalities_2000}, and the intentional binding effect, which utilizes the reported time interval between the action and consequence \cite{haggard_voluntary_2002}. 
Some work in HCI has applied these implicit methods to measure user agency in computer assistance \cite{coyle_i_2012}, but more recent work has questioned whether the results from these techniques are truly predictive of user outcomes \cite{bergstrom_agency_ux_2022}. Explicit methods of measuring sense of agency include self-reflection and questionnaires of the influence of one's actions on an outcome. In our work, we measure sense of agency using a six-item questionnaire inspired by Tapal et al.'s Sense of Agency Scale \cite{tapal_sense_2017}.

\begin{table*}[!bt]
    \centering
   \caption{Robot Autonomy Levels Presented in Survey}
   \begin{tabular}{|c|c|c|c|}
   \hline
   \textbf{Robot Type} & \makecell{\textbf{Is Robot Acting} \\ \textbf{ Autonomously}} & \makecell{\textbf{Who Controls Program} \\ \textbf{(if autonomous)}} & \makecell{\textbf{Who Controls Actions} \\ \textbf{(if not autonomous)}} \\
   \hline
   \hline
   Fully Autonomous & Yes & Third Party (Manufacturer) & N/A \\
   \hline
   End-user Programmed & Yes & User & N/A \\
   \hline
   Third-party Teleoperated & No & N/A & Third Party (Operator) \\
   \hline
   Fully User-controlled & No & N/A & User \\
   \hline
   \end{tabular}
   \label{tab:robot_autonomy}
   \end{table*}

\subsection{Robot Autonomy} 

Automation and its impacts on users has been investigated in multiple domains, including aviation and autonomous vehicles \cite{carsten_control_2012,parasuraman_performance_1993}. This work has resulted in taxonomies that characterize different levels of automation \cite{endsley_here_2017}. It has generally been found that as the level of automation increases, the user's awareness of the technology's state decreases \cite{endsley_here_2017}.

In robotics, these levels of automation have been characterized as ``robot autonomy''. Similar to other domains, robotics researchers have widely observed that a user's feeling of control over a robot decreases with increased robot autonomy \cite{beer_domesticated_2012}. Given the decrease in the user's feelings of control, we are interested in how their sense of agency is impacted by robot autonomy. Collier et al. \cite{collier_sense_2025} show that while higher robot autonomy improves task performance in a robot grasping task, the user's sense of agency decreases. 
Other works highlight user preferences for lower robot autonomy in tasks like caring for self or others in healthcare settings \cite{bhattacharjee_is_2020, olatunji_levels_2022, ranganeni_accessteleopkit_2024, bhattacharjee_community-centered_2019, krishnaswamy_multi-perspective_2018, smarr_domestic_2014}.  

 Several definitions have been proposed for categorizing different levels of robot autonomy. Beer et al. \cite{beer_toward_2014} propose that each level of robot autonomy corresponds to a robot's abilities to sense, plan, and act. They suggest that the task complexity and risks should be considered in determining a robot's level of autonomy. More recently, Kim et al.\cite{kim_taxonomy_2024} introduced a taxonomy that categorizes autonomy into six distinct forms, independent of the task complexity. These forms broadly depend on the degree of user involvement before and during the robot's operation. Important to our interests, Kim et al. \cite{kim_taxonomy_2024} point out that human agency and robot autonomy can be conceptualized as a non-zero-sum trade-off. Giving robots more independence does not necessarily reduce human control and decision-making power. One promising robot interaction that has been shown to give users more control while allowing them to benefit from a robot acting autonomously is end-user programming \cite{ranganeni_accessteleopkit_2024, zhang_social_agency_eup_2025}. An end-user programmed robot acts autonomously in accordance with the user's specified program. This contrasts with a fully autonomous robot, which acts autonomously in accordance with a program created by the manufacturer, over which the user has no control.
 
\subsection{User Preferences for Robot Autonomy}

User preferences for robot autonomy and usage have been investigated in multiple contexts. In robot-assisted feeding and navigation contexts, fully autonomous robots are undesirable as users prefer to be in control at different points of the robot's trajectory \cite{bhattacharjee_community-centered_2019, 10.1145/3568162.3578630}. Across multiple household contexts, older adults prefer robot usage for tasks like chores, information management, and manipulating objects, but do not want robotic assistance for all household tasks \cite{smarr_domestic_2014, beer_domesticated_2012}.

With companies promising to deploy robots in homes \cite{technologies_1x_nodate, Adcock2022FigureAI}, it is important to understand how context impacts the user's desire for control over the robot across a wide array of household scenarios. Our study introduces multiple different scenarios that vary by their risks, which change the user's desire for involvement and control over the robot. We aim to investigate what factors lead a user to prefer robots that increase or decrease their sense of agency across these settings.

\section{USER STUDY}

Although prior work has studied the trade-off between robot autonomy and user's sense of agency for specific robotic systems, most studies limit their comparisons of robot autonomy levels and none have included third-party teleoperated robots. For robot designers to identify the components of a human-robot interaction that most influence a user's sense of agency, we must extend these comparisons. We aim to investigate how users sense of agency is affected by a range of four diverse robot autonomy levels.

\textbf{\textit{RQ1:}} \hspace{0.5em} How does robot autonomy and third party involvement influence a user's sense of agency in using a robot?

Given that higher robot autonomy and third parties controlling the robot decreases the user's control, we propose the following hypothesis:

\textbf{\textit{H1:}} Users' sense of agency will be negatively influenced by the robot acting autonomously and whether a third party is involved in the programming or operation of the robot's actions. 

To test this hypothesis, we will vary whether the robot acts autonomously and
is programmed or operated by a third-party across the four different robots described to the participants (see \autoref{tab:robot_autonomy}). By asking participants to rate their perceived sense of agency for each robot type and provide their reasoning, we will isolate the impact of different factors on the user's sense of agency.

Previous work on the impact of automation on users and their preferences focused on task complexity. Instead, we characterize the tasks by their \textit{risk levels}. We posit that differences in risk scenarios may drive user preferences about need for control of the robot in order to prevent potentially costly consequences.

\textbf{\textit{RQ2:}} \hspace{0.5em} How do users' preferences for robot autonomy change across different household scenarios with varying risks?

Based on the assumption that higher risk leads to a higher user desire for control, we propose to test the following hypothesis:

\textbf{\textit{H2:}} Users will prefer lower robot autonomy and greater control over the robot as task risk increases. 

We hypothesize that greater control will manifest as a desire to preserve their sense of agency by programming or operating the robot's actions themselves. We will test this hypothesis by asking users to rank their preferences for four robot types in eight scenarios with varying risk levels and types of risks. Through these rankings and qualitative analysis of the open-ended answers, we will analyze how the users' preferences vary across tasks and what factors influenced their preferences.

\subsection{Study Design Part 1: Sense of Agency vs. Robot Autonomy}\label{sec: user study part 1}
The first part of the survey asked participants to rate their perceived sense of agency per robot. The robot autonomy levels were varied for each robot by changing whether it acts autonomously during its operation and whether the user or a third party is involved in specifying the robot's program.

First, the Hello Robot Stretch was introduced as the robot platform to align participants’ expectations of the robot's morphology and abilities. Participants were then shown a 90 second video of Stretch operating in real-life household scenarios, adapted from Hello Robot's video.\footnote{\url{https://www.youtube.com/watch?v=Ni4p8axgqHM}} The video aimed to help participants appreciate Stretch’s general capability of completing different tasks but did not specify the robot autonomy level.

The survey presented four different autonomy levels as different Hello Robot Stretch configurations, with \autoref{fig: robot type} showing the images and descriptions provided to participants. As each robot was introduced, participants were shown six five-point Likert scale questions based on Tapal et al.'s Sense of Agency Scale, a validated scale that measures an individual's sense of agency \cite{tapal_sense_2017}. The questions, presented in random order, asked them to rate their perceptions of (1) the degrees of control, (2) intention, (3) predictability of outcomes, (4) free will, (5) decision making, and (6) responsibility 
if they were to hypothetically use the robot. After answering each set of questions, participants answered an open-ended question about the factors that influenced their ratings for each robot. 

Although the four different autonomy levels presented in the survey (fully autonomous, end-user programmed, third-party teleoperated, and user-controlled) do not represent every possible robot control or autonomy level, we chose them as representative of four distinct and prominent robot interaction modalities being developed in industry and academia. They were also chosen because they encompass a wide range of robot autonomy levels and forms as categorized by prior work \cite{beer_toward_2014, kim_taxonomy_2024}.

\subsection{Study Design Part 2: User Preferences for Robot Autonomy vs. Task Risks}

To address the second research question, participants were introduced to eight task scenarios in a randomized order. The scenarios each encompassed a task that varied in risk and goal. We created these tasks to represent multiple types of household activities and needs, inspired by the tasks surveyed in \cite{smarr_domestic_2014}. 

The survey introduced each task to the participants with a detailed description, visual depiction, and title. Two tasks addressed \textit{simple chores}, e.g., assistance with watering plants and bringing in the mail when the user is away from home. Two more tasks entailed \textit{caregiving}, including help with feeding the dog and playing board games with their ill mom while the user is out of town. The remaining two tasks required the robot to \textit{make decisions that may impact the health} of the user or their family. These scenarios included a robot helping prepare an after-school snack for a young child with allergies and helping fetch the user's prescribed medication from the kitchen during their recovery from surgery. We omitted two tasks from our analysis due to participant confusion and differing implicit assumptions about answers. 

After introducing each task, participants rated their assessment of (1) the severity of the task's potential consequences and (2) their desired involvement in completing the task with human or robotic assistance on a five-point Likert scale (1=Very Low, 5=Very High). Participants also ranked the four robot types by dragging and dropping preferences from ``Most Preferred'' to ``Least Preferred.'' When finished, they were asked to complete an open-ended question about their reasoning for the robot rankings on the tasks.

\subsection{Analysis}
We averaged each participant's ratings for the six questions for each robot type on a five-point Likert scale (1=Strongly Disagree, 5=Strongly Agree). Cronbach's $\alpha = .847$. We analyzed the factors that influence a user's sense of agency using a robot through a mixed linear model estimated with Restricted Maximum Likelihood (REML). Each autonomy level (fully autonomous, end-user programmed, third-party teleoperated, and user-controlled) is comprised of a different set of binary variables that identify its unique properties (listed in \autoref{tab:robot_autonomy}). The model examined factors potentially influencing participants' sense of agency when interacting with robots, including the binary variables and themes coded from open-ended responses. Participant Survey ID was included as a random effect to account for the repeated measures design, with each participant providing responses across multiple robot types.

We used an iterative coding process to analyze open-ended responses from both parts of the survey. The first and second authors independently read all qualitative responses and noted different themes. This led to the creation of a codebook, which the authors used to independently annotate each response again, with finalized themes showing an inter-rater reliability score of 0.99. Remaining discrepancies were manually resolved through discussion.

\subsection{Participants}
With IRB approval, we recruited participants (N=21) for our study through convenience sampling at an American university. In the recruitment flyer, participants received a link to the survey and were informed that it would take around 30 minutes to complete both parts. Each participant was reimbursed with a \$12 gift card when they fully completed the survey. 

Eighteen of the participants identified as female and 3 as male. All were between the ages of 18–34. All but one participant disclosed some college education or an advanced degree. The participants rated themselves as generally knowledgeable about and excited to use robots in the future but inexperienced with using robots in their daily lives.  The complete set of study materials, anonymized participant responses, and codings were uploaded to OSF.\footnote{\url{https://osf.io/y5n4t/?view_only=efb15dcc0e444e7f85ce4a940aaf3496}}

\begin{figure}
    \centering
    \includegraphics[width=\linewidth]{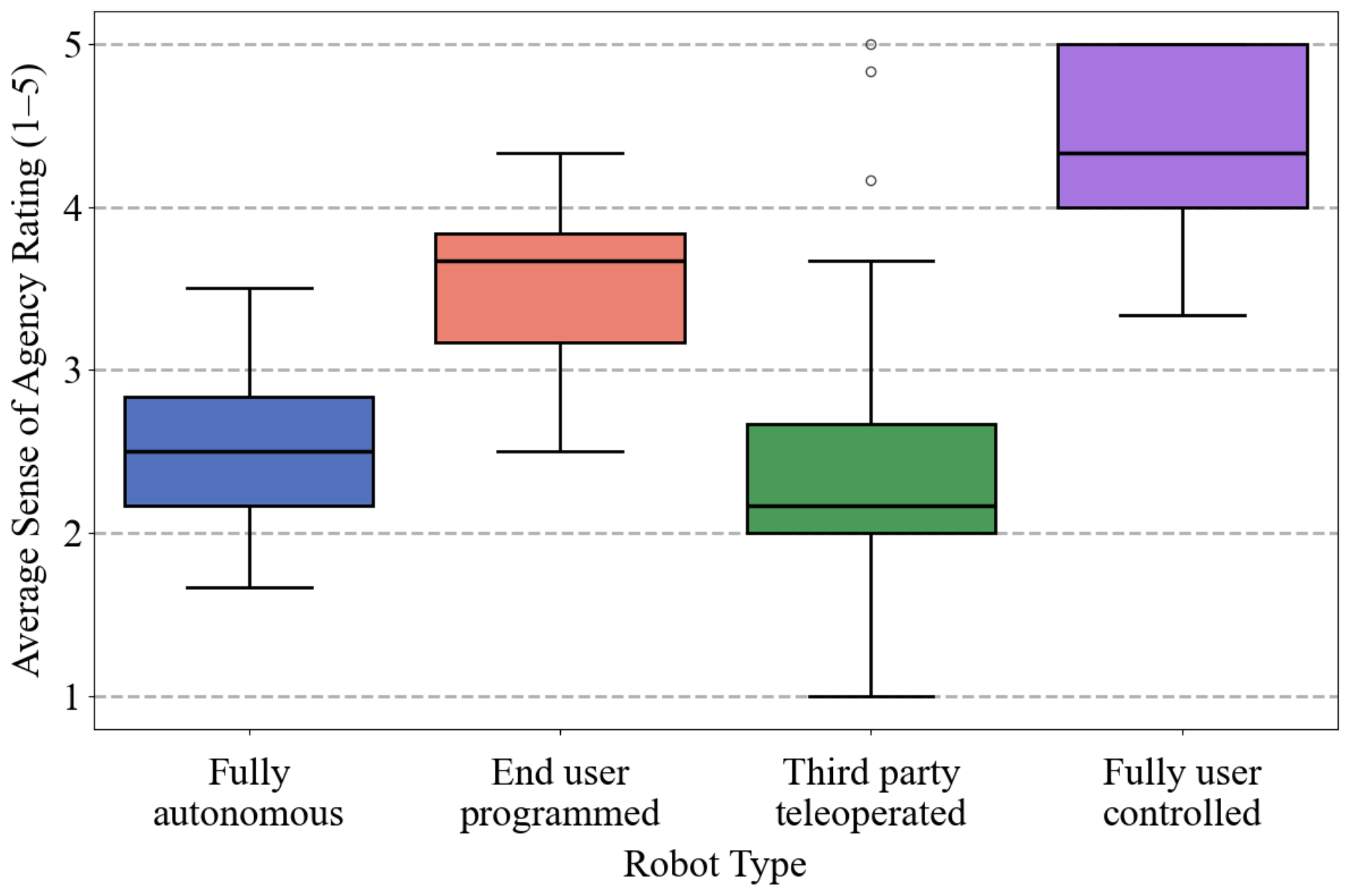}
    \caption{Average Perceived Sense of Agency by Robot Type. Participants rated the fully user-controlled robot and the end-user programmed robot as the robots that would give them the highest and second-highest sense of agency, respectively. There was no significant difference found between the ratings for the fully autonomous robot and third-party teleoperated robot.}
    \label{soa-viz}
\end{figure}

\section{RESULTS}

There were two main findings from this work. First, third-party involvement in the programming or operation of the robot was found to be a highly significant factor in decreasing a user's sense of agency, more so than the robot acting autonomously. We found that user sense of agency is increased by giving the user control over the programming of the robot, rather than giving a third-party operation over the robot. Second, users tended to prioritize preserving their sense of agency more as the task's perceived risk increased. This desire to preserve their sense of agency trades off their preferences for end-user programmed robots vs. third-party teleoperated robots, depending on whether the user trusts third party teleoperators for the task.

\subsection{Factors Influencing Robot User's Sense of Agency}

We find that, on average, participants rated their perceived sense of agency higher for robots they have control over programming or operating (see \autoref{soa-viz}).

\begin{table}[b]
    \centering
   \caption{Factors Influencing Robot User Sense of Agency}
   \begin{tabular}{lcccc}
   \hline
   \textbf{Predictor} & \textbf{$\beta$} & \textbf{SE} & \textbf{z} & \textbf{p} \\
   \hline
   Intercept & 3.365 & 0.138 & 24.412 & $<$.001 \\
   Robot Acts Autonomously & -0.865 & 0.167 & -5.175 & $<$.001 \\
   Third Party Is Involved & -1.996 & 0.175 & -11.392 & $<$.001 \\
   User Trusts Third Party & 1.250 & 0.368 & 3.401 & .001 \\
   Robot Autonomy $\times$ Third Party & 0.980 & 0.242 & 4.048 & $<$.001 \\
   \hline
   \end{tabular}
   \begin{flushleft}
   \small{Note: Random effect variance for Participant Survey ID = 0.106}
   \end{flushleft}
   \label{tab:soa_model_results}
   \end{table}

Analysis revealed that a robot acting autonomously ($\beta = -0.865, p < .001$) and a third party being involved ($\beta = -1.996, p < .001$) significantly decreased the participants' sense of agency in human-robot interactions (see \autoref{tab:soa_model_results}), supporting H1. The interaction between the two variables was also significant ($\beta = 0.980, p < .001$). The magnitude of these coefficients and significance of the interaction indicates that third-party involvement has a substantially larger negative effect on agency than robot autonomy. The random effects variance of 0.1 indicates moderate individual differences in baseline ratings across participants, which is accounted for in the model.

Notably, third-party involvement in the programming or control of the robot most significantly reduced the user's sense of agency, more so than whether the robot acted autonomously. This finding suggests an important design principle:\textbf{ users' sense of agency can be preserved even when interacting with autonomous robots as long as users themselves program the robot.} Our model predicts that an end-user programmed robot would yield a sense of agency rating of 2.5, which is significantly higher than a fully autonomous robot programmed by a third party, with a predicted rating of 1.5.

While the fully user-controlled condition predictably results in the highest sense of agency of 3.4, the end-user programmed robot offers a valuable opportunity to preserve substantial user agency while still offering the benefits of robot autonomy. Even when considering users who trust third parties (a positive factor with $\beta = 1.250, p = .001$), the end-user programmed robot's advantage over the third-party teleoperated option is still present. Assuming that an individual user fully trusts a third party, the predicted sense of agency ratings for the third-party teleoperated robot and end-user programmed robot are roughly equivalent (2.6 and 2.5, respectively). It is unlikely that users 
will fully trust third parties when using robots for all tasks in a household, however. More likely, trust levels will vary depending on the context. 

\subsection{Relationship between Task Risk and Desired Involvement}

\begin{figure}
    \centering
    \includegraphics[width=\linewidth]{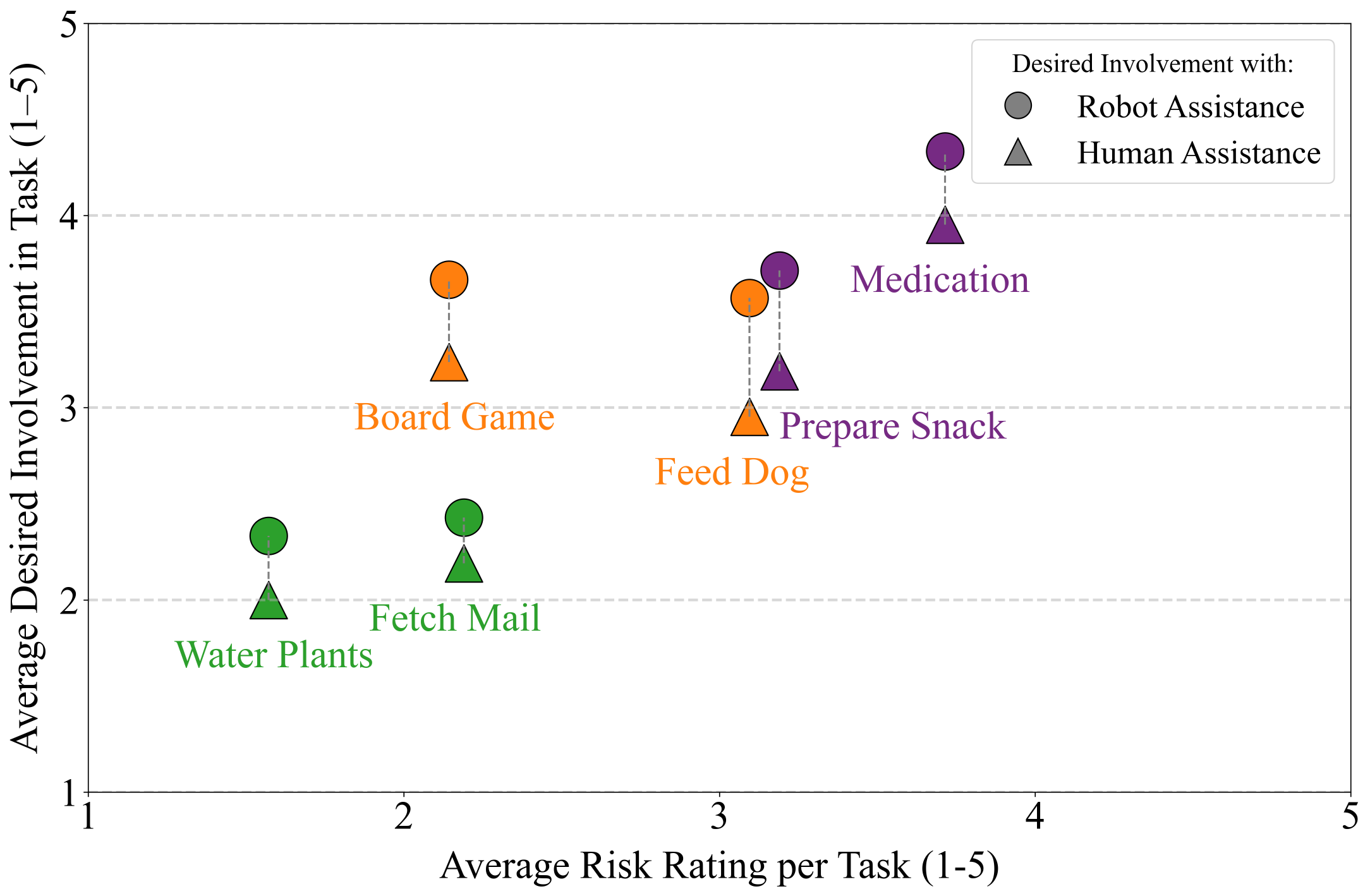} 
    \caption{Participants' Average Ratings of Risk vs. Desired Involvement for Each Task. There is generally a positive relationship between the rating of risk in the task and the desired involvement.} 
    \label{fig: risk_vs_involvement} 
\end{figure}

As expected, respondents' desired involvement in the task increased as their rating of task risk increased. Further, they preferred to be more involved in a task when receiving robot assistance compared to human assistance, as shown in \autoref{fig: risk_vs_involvement}. These generally confirm the notion that users want more involvement and control over robots in riskier situations and provides evidence to support H2. However, risk and desired involvement alone do not account for the variance in the participant's ranked preferences for robot types for each task.
We found that other contextual factors and concerns significantly influenced their preferences, through analyzing the open-ended responses.

\subsection{Factors in User Preferences for Robot Autonomy across Varied Task Risks}
We grouped the tasks by similarity, based on the involvement level and nature of the task's risk (\autoref{fig: risk_vs_involvement}).

\subsubsection{Watering Plants and Fetching the Mail}
For low risk tasks, participants had a lower desire for involvement and preferred autonomous robots, supporting H2. The fully autonomous and end-user programmed robots were the most preferred robots. Over 90\% of participants who emphasized that these tasks were low risk in their open-response ranked the fully autonomous robot or the end-user programmed robot as their first choice, showing that users prefer to use autonomous robots in low risk situations.

Written responses explain why the end-user programmed robot was a popular top choice: ``[Getting the mail] seems like a benign task but still I won't prefer giving up entire autonomy,'' ``[I] can teach it and test it, and also correct it remotely," and ``It can be precisely told what to do.'' This highlights that \textbf{users feel that programming the robot preserves their sense of agency, despite the robot acting autonomously}.

The third-party teleoperated and fully user-controlled robots were most frequently ranked last. Over 33\% of the participants expressed security or trust concerns like: ``[I] don't want a third-party stranger seeing [my] packages\ldots and potentially my address,'' ``They can see sensitive information,'' and ``I don't know what the third-party person could do.'' One participant mentioned ``I would not trust a third-party\ldots unless they too happened to be a plant person,'' suggesting that emotional attachment impacts users' trust in third-party teleoperation. 

We observed some individual variation in these trends. Across these low-risk scenarios, several participants chose the third-party teleoperated robot as their most preferred robot, citing trusting other third party operators for these tasks, desiring a human to be involved, and wanting low self-involvement.

\subsubsection{Playing a Board Game with Mom and Feeding the Dog}
\begin{figure}
    \centering
    \includegraphics[width=\linewidth]{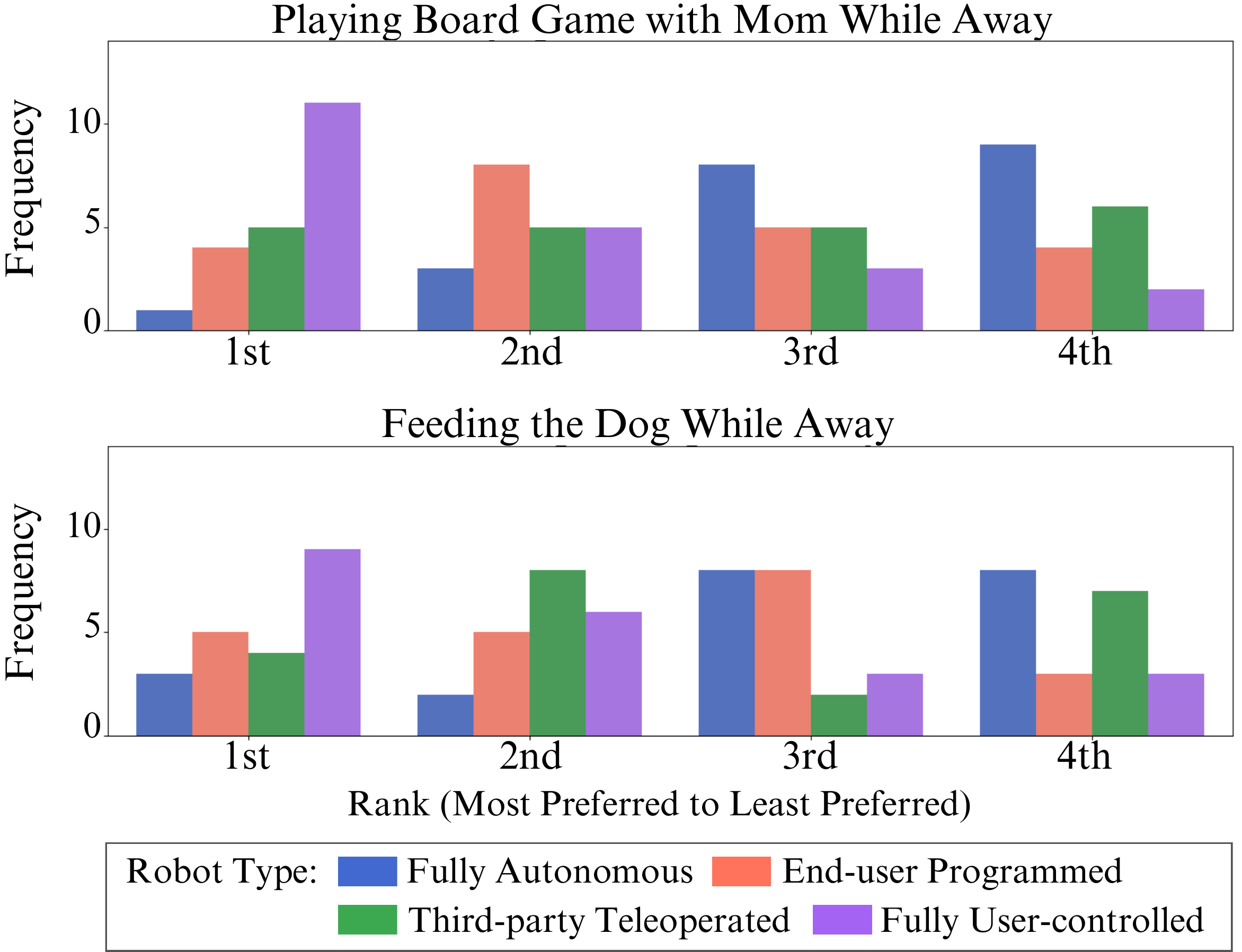}
    \caption{Histogram of ranked robot types for the medium risk tasks, where the robot interacts with the user's loved ones. See OSF for histograms of the other task preferences.}
    \label{fig: ranking_feeding_playgame}
\end{figure}

For tasks where the robot interacts with the users' loved ones, participants most frequently preferred the user-controlled robot the most, but were mixed on whether they preferred a third-party teleoperated robot or an autonomous robot second (see \autoref{fig: ranking_feeding_playgame}), only partially supporting H2. In caregiving scenarios with medium risk, some users may be willing to sacrifice their sense of agency by using the third-party teleoperated robot over the end-user programmed robot. Participants cited social and emotional reasons for wanting a human (even if it were not themselves) to control the robot. In feeding the dog, a participant stated that they prefer either themselves or a third party controlling the robot because ``It would be like hiring a pet sitter to feed your dog''. In playing board games with mom, a participant stated, ``Having more direct control over the robot or having another person control the robot would make me more confident that the elderly mother was feeling less lonely''. 

Despite the overall preference for non-autonomous robots in these situations, 43\% of participants preferred end-user programmed robots over third-party teleoperated robots. The participant's trust in the third-party teleoperator also highly impacted their ranking order.
Among the approximately 20\% of participants who mentioned distrusting third parties, 100\% of them ranked end-user programmed robots over third-party teleoperated robots across both scenarios. 
Participants preferred end-user programmed robot because they wanted the robot to feel familiar in interacting with their loved ones, with one stating, ``I think the important factor is which robot best represents me\ldots [the end-user programmed robot] learns from me directly\ldots [the autonomous robots] wouldn't provide the same representation no matter how well the games are played.". This demonstrates that \textbf{while participants generally preferred non-autonomous robots in caregiving situations, users who distrust third party teleoperators prefer an end-user programmed robot over the third-party teleoperated robot because they believe that its program can represent their intentions.}

\subsubsection{Preparing a Snack for Child with Allergies and Fetching Medications}
For tasks involving high-risk health decisions, users overwhelmingly rank the user-controlled robot first,
supporting H2. 
Half of the participants cited high task risk and wanting a higher sense of agency. Some notable responses include: ``Need agency in these decisions,'' ``I only trust myself here,'' and ``[I] want full control\ldots no matter where I am to ensure it does it right.'' The fully autonomous robot was the last ranked robot across both tasks, showing that users highly prefer using a robot that preserves their sense of agency in high risk situations.

\section{DISCUSSION}

\subsection{End-User Programming for Preserving User Agency in Autonomous Robots}
Participants expressed distrust and concern about the reliability of a fully autonomous robot, but expressed that the robot acting autonomously would be convenient in low risk scenarios. End-user programmed robots address this trade-off by allowing the users to both preserve their sense of agency while benefiting from the convenience of an autonomous robot. This insight provides evidence that counters the widespread notion that a user's control needs to decrease as a robot's autonomy increases \cite{bradshaw_seven_2013}. Our sense of agency factors model (\autoref{tab:soa_model_results}) also supports this, showing that a third party being involved in the programming or control of the robot has a greater negative effect on the user's sense of agency, compared to the robot acting autonomously. User control of the robot's programming enhances the sense of agency by replacing the third party manufacturer as the robot programmer. We recommend further work investigating how end-user programming and other methods can preserve the user's control while the robot acts autonomously.

\subsection{Security and Privacy Concerns Impact Preferences in Caregiving Tasks}
Caregiving tasks that require social/emotional judgments (e.g., playing board game with mom while away, feeding the dog while away) showed a trade-off between allowing the robot to act autonomously (with end-user programming) and a robot not acting autonomously (with a third party operator controlling the robot). Users who trusted the third party operators preferred the third party operator controlling the robot over the end-user programmed robot, and vice versa. Multiple participants cited privacy and security concerns as a significant factor impacting their degree of trust in third party operators. While previous work in HRI has found that people prefer third parties teleoperate robots with filtered videos to protect their privacy \cite{butler_privacy-utility_2015}, it is also notable that security and privacy concerns have been cited as a major barrier to user adoption of in-home Internet of Things devices \cite{schuster_users_2024_fixed}. Given that robotics startups and companies aim to deploy third-party teleoperated robots to the home in the near future \cite{torch_invasion_2025, humans_behind_robots_article}, 
we recommend researchers investigate the privacy and security of teleoperated robots and the impact of these concerns on the adoption of household robots.

\subsection{Additional Concerns Impacting User Preferences}
Participants expressed several other sentiments that also impacted their sense of agency or preferences. They are outside the main focus of this study, but worth mentioning for future investigation: 
\begin{itemize}
    \item Participants stated that their decision to purchase and use the robot increases their sense of agency, since it makes them responsible for the robot even if the robot is acting autonomously or is operated by a third-party.
    \item The third-party robots were ranked lower by some participants for low risk scenarios for fear of wasting the third-party operator's time with non-consequential tasks.
    \item Participants ranked the user-controlled robot lower in their preferences because they were concerned about their own ability to control it effectively.
    \item Assumptions surrounding the availability of features like video streaming and voice commands improved the ranking of autonomous robots as participants stated these features would make them feel reassured of the robot's reliability. 
\end{itemize}

\section{LIMITATIONS AND FUTURE WORK}
Our study's limited sample size and recruitment pool may not fully represent the views of the general public, as age and education level may have a large influence on participants' responses. The vignette-based survey method also limited participants to respond based on descriptions, rather than through experience using a robotic system. Future work should use a larger, more demographically diverse sample and extend these findings through studies with implemented/deployed robotic systems. Such studies could also investigate how user's preferences for robot autonomy change with increased familiarity and usage.

Several promising directions for future research emerge from this work. First, future work could expand the set of autonomy levels to investigate shared autonomy and variable autonomy robot interactions. In shared autonomy, the user's degree of control over the robot changes throughout the robot's operation. In variable autonomy, the robot's level of autonomy varies based on the changes in the context \cite{reinmund_variable_2024}. Because these interactions dynamically trade off the user's control of the robot, they warrant further research on how they impact the user's sense of agency. Investigating these approaches could inform the design of human-robot interactions that enhance the user's sense of agency by contextually changing the robot autonomy in accordance with the user's preferences. Second, it is important to work with physically disabled populations to understand how their perceptions of agency and preferences differ across robot autonomy levels and contexts. As perceived control of the robot has been shown to have a significant effect on the intention of these populations to use care robots \cite{jung_factors_2024}, this research is crucial towards informing the design of these robots intended to assist users with activities of daily living.

\bibliographystyle{IEEEtran} 
\bibliography{references.bib,references_editable.bib}

\addtolength{\textheight}{-12cm}

\end{document}